**Face processing emerges from object-trained convolutional neural networks**

Zhenhua Zhao[1¶], Ji Chen[2¶], Zhicheng Lin[3*], & Haojiang Ying[1*]

[1.] Department of Psychology, Soochow University, Suzhou, China

[2.] Center for Brain Health and Brain Technology, Global Institute of Future Technology,

Institute of Psychology and Behavioral Science,

Shanghai Jiao Tong University, Shanghai, China

[3.] Department of Psychology, University of Science and Technology of China, Hefei, China

*Correspondence:

Dr. Zhicheng Lin, zhichenglin@gmail.com

*and*

Dr. Haojiang Ying, hjying@suda.edu.cn

¶ These authors contributed equally to this work.

Data, analysis code and materials can be found on (https://osf.io/kta5v/?view_only=33cc79e205dd4a048b19254bbc7164b7)

# Abstract

The origin of face processing capabilities—whether they arise from specialized neural mechanisms or general object recognition systems—has long been debated. Direct investigation in humans is complicated by lifelong exposure to both faces and objects. Recent advances in convolutional neural networks (CNNs) allow us to test these competing hypotheses by controlling initial training exposure. Domain-general theories predict that face processing can emerge from object recognition mechanisms, while domain-specific accounts suggest that dedicated face processing architecture is necessary. We tested these predictions using CNNs pretrained exclusively on object recognition. Through transfer learning that modified only the final fully connected layer, these networks were tuned to either face–object categorization or 1000-class face identity recognition tasks. The networks demonstrated human-like performance patterns, including face-specific inversion effects and sensitivity to face-like configurations in pareidolia images. These effects persisted even when animal images were excluded from pretraining, demonstrating that face processing capabilities emerge independently of exposure to animal faces. Analysis of network representations revealed both specialized and shared processing units for faces and objects in the final layer, with faces showing distinct activation patterns. These results demonstrate that sophisticated face processing, including identity recognition, can emerge from general object recognition architectures without face-specific pretraining. Our findings support domain-general theories of face processing and demonstrate how CNNs can illuminate fundamental questions about human visual cognition.

**Author Summary**

This study examines whether convolutional neural networks (CNNs), originally trained exclusively on objects, can adapt to identify faces—a task traditionally believed to require specialized neural mechanisms. The findings reveal that general visual processing systems, even without prior exposure to human faces, can develop the ability to recognize faces and replicate human-like perceptual phenomena, such as the face inversion effect. These results challenge the notion of a strictly dedicated origin for face recognition and instead suggest that it may emerge from the broader architecture of visual processing systems. Further, this work highlights the value of CNNs as models for understanding the computational principles underlying visual processing.

**Keywords:** Convolutional neural networks (CNNs), face identity recognition, transfer learning, object representation, face pareidolia

# Introduction

Face recognition is a fundamental cognitive function of the human visual system [1–4]. Numerous studies have identified brain regions and circuits involved in face processing, particularly the fusiform face area (FFA). The right FFA has been found to play a crucial role in face recognition [5–7] such that it exhibits increased activity in response to faces than other objects [8,9], and lesions to this region lead to selective deficits in face recognition [10,11].

However, the FFA (and other face-sensitive regions) is not exclusively selective for faces. For instance, it can also respond to images of houses [12], Greebles (a kind of artificial non-face stimuli) [13,14], and even radiological images such as X-rays, which bear no resemblance to faces [15,16]. Indeed, contrary to the modular (domain-specific) hypothesis [17,18], the expertise hypothesis posits that the FFA is not specific to faces but rather supports specific cognitive processes [19,20]. In particular, accumulating evidence suggests that the FFA supports automatic processing of subordinate-level visual information driven by expertise, extending beyond face perception per se [21–23]. Consequently, the FFA's responses may primarily depend on the physical properties of the objects, such as visually homogeneous categories [24–26]. Therefore, according to the expertise hypothesis, while the FFA is essential for face identity recognition, its representation extends beyond basic image processing and encompasses higher-level perceptual and social aspects of faces [7].

A direct test of the domain-specific hypothesis would require recruiting individuals who have never encountered human faces but are experienced with objects. This approach is grounded in the idea that domain-specific visual experiences are critical for developing specific processing modules [27]. However, recruiting participants with only prior experiences with objects but not human faces is nearly impossible.

Recent advances in computer science have facilitated the simulation of the human visual system using neural networks [28–30]. In particular, convolutional neural networks (CNNs)—comprising a series of convolutional, pooling, and fully-connected (FC) layers—have achieved state-of-the-art, human-level performance in various complex visual tasks, including object categorization [31] and gender classification [32]. The convolutional layers of deep CNNs, which apply a set of learnable filters (kernels) to the input to extract local features, have been found to bear a strong resemblance to the human visual system at both perceptual and higher processing levels [33–36]. Indeed, several studies have suggested that CNNs may utilize similar representations to humans for face identity recognition and gender classification [37–39]. Thus, CNNs offer a viable alternative to human participants that is uniquely suited to address the role of experience in face perception, as they allow precise control over their prior experience (training).

To utilize CNNs for understanding the computational mechanisms of face processing, it is essential to determine what visual information is extracted from face images [39,40]. This requires testing CNNs that demonstrate a certain level of face processing capability. However, many studies analyze activation patterns of CNNs when viewing face images, often relying solely on pre-trained weights from object recognition tasks [41,42]. While these CNNs exhibit *activation* in response to face images, these patterns may not necessarily reflect genuine face *processing*. In this study, we used CNNs pre-trained on objects. We fine-tuned them on either binary transfer task involving "object–face" categorization (Experiments 1 & 2), or a 1000-class face identity recognition task (Experiment 3). This approach enables us to evaluate whether object-pretrained CNNs can develop the capacity to *process* faces beyond mere *activation* patterns.

We conducted three experiments to investigate whether face processing capabilities can emerge from object-pretrained neural networks. In Experiment 1, we examined whether object-pretrained CNNs could categorize, represent, and process faces in a manner similar to humans. Experiment 2 extended this investigation by testing models trained on datasets where “animal faces” were further excluded. Finally, Experiment 3 directly assessed the performance of these networks on a transfer task involving face identity recognition.

To test whether object-pretrained CNNs can develop face processing abilities, we examined their ability to differentiate between objects and faces and assessed whether they exhibited hallmark face effects, such as the face inversion effect and face pareidolia effects. The face inversion effect refers to the disproportionate drop in identity recognition of upside-down (inverted) faces relative to upright ones [43,44]. Face pareidolia involves the perception of faces in images that do not contain actual faces [45]. For instance, when food is arranged to resemble facial features/structures, individuals often report perceiving “faces” [46]. This peculiar phenomenon bridges “face-object” categorization by demonstrating how our brain interprets non-face stimuli as faces. Importantly, it can be harnessed to probe “face” representation in CNNs: if object-pretrained CNNs can develop face representations, they should also be susceptible to face pareidolia. Therefore, analyzing the ability of CNNs to identify “faces” in pareidolia images offers a rare window into the presence and characteristics of “face” representations in these networks.

# Experiment 1

In Experiment 1 (see Fig 1 for an overview of the core method), we used five CNNs initialized with pre-trained weights for object recognition to simulate human participants without exposure to human faces. We then applied a transfer learning task to examine their capability to detect, represent, and process faces. To compare face processing between CNNs and humans, we evaluated their sensitivity to inverted face images and face pareidolia images. Furthermore, to understand how the last FC layer—where the highest level of representation is processed [47–50]—differentiates faces from objects, we examined the connection between the activation levels of its units and the model's success in identifying faces and objects.

**Fig 1 | Overview of the experimental design across three experiments.** CNNs (VGG11, VGG13, VGG16, ResNet101, and DenseNet169) were pre-trained using object recognition weights from either ImageNet-1K (Experiment 1) or Caltech-256 with and without animal images (Experiment 2). In Experiments 1 and 2, these networks underwent transfer learning on an object versus face binary categorization task. The models were then tested with face pareidolia images to evaluate face-processing capabilities, and with upright and inverted images to assess inversion effects. Tensor data from the fully connected layer was extracted to analyze internal representations of faces versus objects. In Experiment 3, the same pre-trained networks were evaluated on a 1000-class face identity recognition task. Pareidolia images were obtained from a publicly available Flickr.com image pool (https://www.flickr.com/groups/facesinplaces/pool/); icons are used with permission from www.flaticon.com.

# Methods

## CNN setups

We initialized the CNN with ImageNet-1K dataset weights from PyTorch to create an object-recognition-only model, excluding face recognition (for details see https://pytorch.org/vision/stable/models.html?highlight=weights). The ImageNet-1K dataset includes 1,000 categories of labeled objects—a diverse range of objects, animals, and scenes but no human faces (all human faces were masked and blurred, https://image-net.org/update-mar-11-2021.php; for additional details see [51]). The last layer of the CNNs (a SoftMax layer) outputs a probability distribution over the potential categories (e.g., faces vs. objects). The CNNs, equipped with these specific weights, are therefore naïve in representing human faces. To ensure generalizability, a total of five CNNs were examined, including VGG11, VGG13, VGG16 [52], ResNet101 [53], and DenseNet169 [54], as detailed in Supplementary Materials.

## Transfer learning

To assess the ability of CNNs to recognize faces, we first conducted transfer learning, where CNNs were tasked with categorizing faces versus objects based on the content of the images using generic visual categorization [55]. Similar to [40], we used the cross-entropy function from PyTorch to compute loss, and the Adam optimizer (with a learning rate of 0.0001) to optimize trainable parameters. Within each epoch, parameters were adjusted during the training phase and then evaluated during the validation phase. If these adjustments improved validation accuracy, surpassing the highest accuracy achieved in prior epochs, the parameters were then preserved. Throughout this process, all network parameters were frozen except for the final FC layer, thus allowing only the use of pre-trained object weights to represent faces.

There were 40 epochs (i.e., rounds) of transfer learning, each including a training phrase followed by a validation phase. To mitigate overfitting, the two phases used different sets of images. The face images were sourced from the Wider Face dataset [56], while the object images came from the Caltech-256 dataset [57], excluding any faces or humanoid figures. To minimize bias, the image quantities were balanced: 12,880 face images and 12,658 object images for training; 3,226 face images and 3,177 object images for validation. All images were standardized to 224 × 224 pixels, and image RGBs were normalized into values between 0 and 1, with a mean of 0.5 and a standard deviation of 0.5. Thus, categorization could not be based on identity-irrelevant systematic differences between the two datasets. Categorization accuracy was recorded.

Furthermore, the tensors entering the fully connected layer—the CNN's highest task-specific representations—were extracted to further compare potential similarities and differences between face and object representations through representational similarity analysis, as detailed below in the “object and face processing units” section.

## Tests on pareidolia images and inversion effect

Following transfer learning, we further assessed face processing mechanisms in the CNNs by testing their performance with face pareidolia and inverted face images. This assessment evaluated whether the CNNs identified faces in pareidolia images at a higher probability than the base rate for mistaking regular object images as faces. To establish this baseline and assess categorization robustness, the CNNs were presented with 300 object images not encountered during the transfer learning phrase. These images—comprising 10 categories with 30 randomly selected labeled images per category—were drawn from the STL-10 dataset [58].

For the pareidolia test, 300 face pareidolia images were selected from the Faces in Places Flickr group (https://www.flickr.com/groups/facesinplaces/pool/). To ensure that these images indeed looked like faces, we recruited 30 human participants to evaluate each image for its resemblance to a face through the procedure shown in Fig 2. The sample size was determined based on a recent study with a similar experimental design [40]. Participants, naive to the purpose of the experiment, were recruited from Soochow University and had normal or corrected-to-normal vision. [**Ethics Statement:** This study was approved by the Institutional Review Board (IRB) at Approval of Ethics committee of Soochow University (Approval No. SUDA20210402H03). All participants provided written informed consent prior to participation. The experiment was conducted in accordance with the Declaration of Helsinki and all ethical guidelines for research involving human subjects. Participants were informed of their right to withdraw from the study at any time without consequence, and no personally identifiable information was collected. Data were anonymized and securely stored to ensure participant confidentiality. All procedures were designed to minimize any potential risks to participants, and appropriate measures were taken to ensure their comfort and well-being throughout the study. ]

**Fig 2 | The trial sequence for the face pareidolia experiment with human participants.** Participants fixated on a central cross, which disappeared after 500 ms, followed by a face pareidolia image displayed for 500 ms. Subsequently, two red-framed rectangles appeared, and participants used a mouse to click on either the "Yes" or "No" region to indicate whether the image looked like a face.

To determine if the CNNs exhibited the classic face inversion effect—characterized by higher recognition probabilities for upright than inverted faces [44,59]—we tested them with 300 new face images featuring identities not included in the transfer learning dataset. These images were sourced from the VGGFace-2 dataset [60]. For comparison and as a baseline, we also examined the inversion effect on objects by testing the CNNs with 300 inverted object images from the STL-10 dataset [58].

## Object and face processing units

To study how units in the final FC layer differentiated faces from objects, we analyzed the correlations between unit activations and accurate face and object categorizations. Using "forward hooks" [61–64], we captured tensors at the entry point of the last FC layer while exposing CNNs to 300 face images from VGGFace-2 [60], 300 face pareidolia images from the pareidolia test, and 300 object images from Caltech-256 [57]. Person correlations were computed for each unit's tensors against the correct categorization outputs, with a significance threshold of $p < 0.05$ applied to identify units potentially involved in processing specific image types.

Next, we sought to delineate representational similarities among faces, face pareidolia images, and objects. This analysis involved calculating the Euclidean distance between the average tensors of the three image types within object units, face units, overlap units (i.e., units sensitive to both faces and objects), and all units collectively. Confidence intervals were estimated through bootstrapping [65].

# Results

## Transfer learning

Despite architectural differences and the use of only object weights, all the CNNs excelled in discriminating faces from objects after just a few training epochs with feedback. Fig 3 summarizes the results of the transfer learning: after 40 training epochs, the five object-trained CNNs accurately (validation accuracy > .97) categorized faces versus objects—matched for image resolution and color—both for the training images (Fige 3a) and for the new, validation images with no feedback (Fig 3c), indicating robust learning and generalization. The loss, representing prediction error, decreased sharply and stabilized (Fig 3b and Fig 3d), indicating effective learning and convergence towards minimal error rates.

**Fig 3 | Training and validation performance of five CNN models over 40 epochs**. Accuracy (left) remained high while loss (right) was low during training (a and b) and validation (c and d). Both accuracy and loss metrics plateaued within a few epochs, indicating rapid convergence in model performance as training progressed.

## Pareidolia test and inversion effect

Successful categorization of faces versus objects, however, may not necessarily indicate a genuine face processing capacity: the networks could simply rely on broader object versus non-object differentiation. To target face processing capacity, we tested two canonical face-specific effects: pareidolia and inversion effects.

All five CNN models (**Table 1**) exhibited a strong tendency to categorize pareidolia images (object images with face-like patterns) as faces. This categorization occurred at rates much higher than the baseline false alarm rate of categorizing regular object images as faces. Among the models, DenseNet169 showed the highest susceptibility to face pareidolia (20%), followed by ResNet101, while VGG16 and VGG13 exhibited slightly weaker tendencies (13%; main effect of model type: $F(4,1495) = 4.34$, $p = .002$, partial $\eta^2 = .01$).

**Table 1.**

*Probability of categorizing face pareidolia images as faces in CNNs, compared with false alarm baselines (categorizing regular object images as faces during the transfer task).*

| CNN | Mean (%) | Baseline (%) | Pareidolia effect (%, mean – baseline) | $t(299)$ | $p$ | Cohen's $d$ |
|---|---|---|---|---|---|---|
| DenseNet169 | 20 | 6 | 14 | 13.10 | < .001 | 0.25 |
| ResNet101 | 15 | 5 | 10 | 11.77 | < .001 | 0.21 |
| VGG16 | 13 | 5 | 8 | 10.98 | < .001 | 0.18 |
| VGG13 | 13 | 6 | 7 | 9.71 | < .001 | 0.20 |
| VGG11 | 15 | 7 | 8 | 10.15 | < .001 | 0.21 |

Having demonstrated model sensitivities to face pareidolia images, we next examined the inversion effect, a hallmark of face perception. When categorizing inverted face images, all five CNN models (**Table 2**) showed significant decreases in accuracy compared to upright face images (all *p*s < .001). These "inversion effects" emerged even though the face images were not encountered during the transfer learning or pretraining stages. The observed effect sizes, while slightly smaller than human inversion effects ($\eta_p^2 = .29$ in Ashworth et al., 2008 [66]; Cohen's $d = .24$, .31, and .37 in Ge et al., 2006 [67]; $\eta^2 = .37$ in Gerlach et al., 2023 [68]), suggest that the CNN developed representations fine-tuned for canonical, upright face configurations. In contrast, apart from DenseNet169, the CNNs did not exhibit an object inversion effect, indicating a level of face-specific processing similar to that observed in humans.

**Table 2.**

*Face and object inversion effects in CNNs.*

| CNN | Face inversion effect (%, upright – inverted) | *T* test | Object inversion effect (%, upright – inverted) | *T* test | Difference (%, face inversion – object inversion) | *T* test |
|---|---|---|---|---|---|---|
| VGG11 | **8.1** | **$t(366.78) = 6.43, p < .001, d = 0.15$** | -0.8 | $t(598) = -0.47, p = .64, d = 0.21$ | **8.9** | **$t(598) = 4.19, p < .001, d = 0.26$** |
| VGG13 | **5.9** | **$t(361.68) = 5.19, p < .001, d = 0.14$** | 0.0 | $t(598) = -0.004, p = 1, d = 0.20$ | **5.9** | **$t(598) = 3.04, p = .002, d = 0.24$** |
| VGG16 | **5.0** | **$t(320.46) = 4.78, p < .001, d = 0.13$** | -1.9 | $t(598) = -1.32, p = .19, d = 0.17$ | **6.9** | **$t(598) = 3.86, p < .001, d = 0.22$** |
| ResNet101 | **7.6** | **$t(359.63) = 5.56, p < .001, d = 0.17$** | **-3.1** | **$t(577.61) = -2.03, p = .04, d = 0.19$** | **10.8** | **$t(598) = 5.15, p < .001, d = 0.26$** |
| DenseNet169 | **7.7** | **$t(326.67) = 6.29, p < .001, d = 0.15$** | **-3.8** | **$t(589.16) = -2.01, p = .04, d = 0.23$** | **11.6** | **$t(598) = 5.08, p < .001, d = 0.28$** |

*Note*. significant $p$ values were marked in bold.

## Object and face processing units in the final FC layer

The pareidolia and inversion effects collectively demonstrated that while CNN models pretrained with objects-only weights to differentiate faces from objects, their internal representations became attuned to the configural properties that characterize upright, face-like visual patterns during transfer learning.

To investigate the emergence of face and object representations in CNN models, we analyzed the activation patterns in the final FC layer in response to the new face and object images (300 each). As Fig 4 shows, across all five CNN architectures, a subset of units (10–20%) exhibited preferential responses to the face images, with their activations strongly correlating with accurate face categorization ($p < 0.05$). Another subset of units (20–30%) appeared attuned to object properties, as their activation profiles tracked correct object categorization. Additionally, a third subset of units (less than 10%) showed correlations with both face and object predictions, suggesting that these units encoded visual features relevant to both categories.

**Fig 4 | Correlation analysis of unit activation with face and object categorization across the five CNNs.** Unit activations (tensors) were obtained from the final FC layer in response to face, object, and pareidolia images, showing the proportion of units attuned to faces, objects, or both categories.

These correlation maps reveal that CNN representations fractionate into distinct face and object channels after transfer learning. While originally trained only for binary object/face categorization, the networks spontaneously developed subpopulations of units finely tuned to the configural information from faces, distinguishing upright face patterns from other object categories.

## Representational distance among faces, pareidolia, and objects

To investigate how representations of faces, pareidolia images, and objects were organized within the CNNs, we quantified the representational distances between their activation tensors in the last FC layer (**Table 3**). Across all models, the object–pareidolia distance was significantly shorter than the object–face distances, indicating that pareidolia images were represented more similarly to objects than to real faces. This pattern aligns with model performance in illusory face perception, where the majority of pareidolia images were categorized as objects. While the magnitude of distances varied across models, the overall trends were consistent, suggesting that despite architectural differences, the CNNs derived qualitatively similar internal representational geometries turned to face, object, and pareidolia properties through the shared discrimination objective. Moreover, the overlap units across all models effectively distinguished between faces and objects, as indicated by statistically significant tensor distances. These results underscore the ability of CNNs to develop separable representations for faces and objects, even in a transfer learning framework with a simple binary categorization task.

These distance analyses reinforce the presence of distinct face and object units within the CNN representations and reveal a systematic relationship in how real faces and pareidolia images are encoded. The findings show that transfer learning can spontaneously yield representations exhibiting face-specialized response profiles akin to those observed in the primate visual system.

**Table 3.**

*Tensor distances between face, pareidolia images, and objects in the last FC layer.*

| CNN | Comparison | All units | | Overlap units | |
|---|---|---|---|---|---|
| | | 95% CIs | Distance | 95% CIs | Distance |
| VGG11 | Face vs. pareidolia | [17.44, 18.38] | 17.79 | [6.46, 6.99] | 4.43* |
| | Object vs. pareidolia | [7.68, 8.97] | 7.94 | [2.48, 3.23] | 1.84* |
| | Face vs. object | [17.61, 18.57] | 17.95 | [6.62, 7.16] | 4.37* |
| VGG13 | Face vs. pareidolia | [18.08, 19.08] | 18.46 | [6.19, 6.67] | 3.94* |
| | Object vs. pareidolia | [7.72, 9.08] | 7.97 | [2.48, 3.26] | 1.66* |
| | Face vs. object | [18.02, 19.03] | 18.36 | [6.31, 6.78] | 3.83* |
| VGG16 | Face vs. pareidolia | [17.27, 18.17] | 17.60 | [5.41, 5.83] | 4.20* |
| | Object vs. pareidolia | [7.68, 9.02] | 7.93 | [2.19, 2.90] | 1.68* |
| | Face vs. object | [17.00, 17.93] | 17.30 | [5.25, 5.63] | 3.99* |
| ResNet101 | Face vs. pareidolia | [12.69, 13.38] | 12.96 | [3.14, 3.54] | 2.22* |
| | Object vs. pareidolia | [5.54, 6.44] | 5.77 | [2.09, 2.62] | 0.98* |
| | Face vs. object | [11.36, 11.94] | 11.56 | [2.45, 2.65] | 2.10* |
| DenseNet169 | Face vs. pareidolia | [12.72, 13.42] | 12.97 | [2.92, 3.30] | 0.30* |
| | Object vs. pareidolia | [5.46, 6.27] | 5.54 | [1.66, 2.17] | 0.11* |
| | Face vs. object | [11.92, 12.60] | 12.15 | [2.52, 2.77] | 0.28* |
| | | Face units | | Object units | |
| | | 95% CIs | Distance | 95% CIs | Distance |
| VGG11 | Face vs. pareidolia | [9.76, 10.51] | 7.91* | [10.77, 11.37] | 10.41* |
| | Object vs. pareidolia | [3.55, 4.31] | 3.43* | [4.71, 5.78] | 4.44* |
| | Face vs. object | [9.70, 10.45] | 7.73* | [11.27, 11.97] | 10.40* |
| VGG13 | Face vs. pareidolia | [9.71, 10.46] | 7.28* | [10.33, 10.91] | 8.94* |
| | Object vs. pareidolia | [3.60, 4.37] | 3.32* | [4.69, 5.79] | 3.89* |
| | Face vs. object | [9.65, 10.43] | 7.06* | [10.61, 11.26] | 8.69* |
| VGG16 | Face vs. pareidolia | [9.07, 9.73] | 7.36* | [9.16, 9.68] | 9.06* |
| | Object vs. pareidolia | [3.38, 4.12] | 3.20* | [4.28, 5.25] | 3.88* |
| | Face vs. object | [8.81, 9.48] | 7.13* | [9.30, 9.87] | 8.67* |
| ResNet101 | Face vs. pareidolia | [4.39, 4.76] | 3.99* | [6.79, 7.26] | 5.83* |
| | Object vs. pareidolia | [2.46, 3.00] | 1.87* | [3.45, 4.17] | 2.79* |
| | Face vs. object | [3.68, 3.89] | 3.49* | [5.75, 6.06] | 5.26* |
| DenseNet169 | Face vs. pareidolia | [5.51, 5.91] | 0.49* | [5.83, 6.24] | 1.07* |
| | Object vs. pareidolia | [2.27, 2.78] | 0.21* | [3.06, 3.67] | 0.49* |
| | Face vs. object | [5.01, 5.38] | 0.38* | [5.18, 5.50] | 0.87* |

*Note.* CIs = 95% confident intervals; *$p < .05$

# Experiment 2

Experiment 1 indicated that object-pretrained CNNs, following transfer learning (categorizing images as objects or faces), exhibited face perception capabilities. Specifically, the observed inversion effects and the magnitude of the pareidolia “illusions” in the five object-pretrained CNNs closely resembled those documented in humans. Furthermore, representational distance analyses revealed that the object-pretrained CNNs developed “face neurons/units” that specifically activated for face images, as well as “overlapped neuron/units” capable of representing both face and object stimuli.

However, while the use of the ImageNet-1K dataset for pretraining excluded human faces, it did include animal faces. Although animal faces are less effective than human faces in activating the FFA [69], they nevertheless contain facial features that may, for instance, contribute to the emergence of pareidolia perception [70]. Thus, to directly test the effect of animal images, Experiment 2 compared the performance of CNNs with pre-training datasets that either included or excluded animal images.

# Methods

Experiment 2 used the Caltech-256 dataset [57] for pretraining, which consists of 256 object categories, including images from both humans and animals. To exclude human images, eight research assistants reviewed the dataset and identified 12 human-related categories, leaving 244 categories. Additionally, 48 categories containing animal images were identified. In other categories, some images also incidentally included humans or animals, such as vehicle-related categories containing human figures. These human images were excluded, and the animal images were identified.

To examine the effects of animal images on training outcomes, CNNs were pretrained on two versions of the dataset: one excluding only human images and another excluding both human and animal images. In light of previous studies using the larger and more complex ImageNet-21K dataset, which demonstrated good performance after 80 epochs of training [71], we tailored our approach to fit our hardware capabilities. Accordingly, we trained our model for 60 epochs with a learning rate of 0.0001 and a batch size of 32.

# Results

## Pretraining

As shown in Fig 5, training and validation accuracy increased consistently over the 60 epochs across all five CNN models for both datasets. The final accuracy levels for the models pretrained with animal images included were comparable to those trained with animal images excluded, with an average difference of 4% across models. This indicates that the presence of animal-related features in the pretraining data did not confer a significant performance advantage. Similarly, training and validation loss decreased steadily over time, with no notable differences observed between the two datasets. The loss trajectories followed similar patterns regardless of whether animal images were included, further supporting the limited impact of animal-related features on the learning process.

**Fig 5 | Performance in pretraining across the five CNN models over 60 epochs.** Training and validation accuracy increased steadily, and training and validation loss decreased consistently for both datasets. (a) When pretrained on the dataset including animal images and classes, all models exhibited gradual improvement in accuracy and reduction in loss over epochs. (b) When pretrained on the dataset excluding animal images and classes, the models showed similar trends in both accuracy and loss, with comparable final performance to the models pretrained on the dataset including animal images. Despite the difference in dataset composition, the five models (VGG11, VGG13, VGG16, ResNet101, and DenseNet169) achieved nearly identical performance, suggesting that the presence or absence of animal-related categories did not significantly influence training or validation outcomes.

These patterns were observed consistently across all five CNN architectures—VGG11, VGG13, VGG16, ResNet101, and DenseNet169. The comparable performance between the two pretraining datasets demonstrates the robustness of these models in learning generalizable representations. These findings suggest that CNNs rely predominantly on general object features rather than specific animal-related features during pretraining.

## Transfer learning

**Fig 6** illustrates the results of the transfer learning analysis, where pretrained CNN models were evaluated on a binary face-versus-object categorization task. After 40 training epochs (batch size = 32, learning rate = 0.0001), both groups of models demonstrated comparable, high performance, with a minimal difference in mean accuracy of 3% between animal-included and animal-excluded pretraining conditions.

**Fig 6 | Performance in the transfer task across the five CNN models over 40 epochs**. (a) Performance when models were pretrained on the Caltech-256 dataset including animal images. (b) Performance when models were pretrained without animal images. The inclusion or exclusion of animal images in the pretraining dataset did not impact performance during the transfer task, as all five CNN models (VGG11, VGG13, VGG16, ResNet101, and DenseNet169) achieved comparable accuracy and loss metrics in both conditions. Furthermore, both training and validation metrics plateaued after only a few epochs, demonstrating rapid convergence in model performance as learning progressed.

The consistency in performance across the five CNN architectures (VGG11, VGG13, VGG16, ResNet101, and DenseNet169) indicates that the inclusion of animal images in the pretraining dataset did not significantly influence the ability of the models to distinguish faces from objects during transfer learning. Training and validation accuracy plateaued early in both conditions, reaching comparable final performance levels. This rapid convergence suggests that the networks effectively leveraged their pretrained representations to adapt to the new task, regardless of the presence of animal-related features during pretraining.

These findings highlight the robustness of transfer learning in these CNN architectures and suggest that the inclusion or exclusion of animal images does not substantially impact their capacity to learn and generalize to the face-versus-object categorization task. The ability to achieve high accuracy with minimal variation further supports the hypothesis that CNNs derive abstract, transferable features during pretraining that are effective for downstream tasks, even when animal images are excluded.

## Pareidolia test and inversion effect

Overall, the presence of animal images in the pretraining datasets enhanced the pareidolia effect (main effect of training condition, $F(1, 2990) = 5.67$, $p = .02$, partial $\eta^2 = .002$). Additionally, some models were more susceptible to the pareidolia effect than others (main effect of model type, $F(4, 2990) = 6.41$, $p < .001$, partial $\eta^2 = .01$), with varying sensitivity to animal images (interaction of training condition and model type, $F(4, 2990) =$

3.20, $p$ = .01, partial $\eta^2$ = .004).

Specifically, similar to Experiment 1, CNNs trained on datasets that included animal images showed a significant pareidolia effect, with higher probabilities of categorizing pareidolia images as faces compared to the baseline (see **Table 4**). CNNs pretrained on datasets without animal images also demonstrated a pareidolia effect, except for two models (VGG11 and VGG16) showing diminished and nonsignificant pareidolia effects when trained on datasets excluding animal images. Simple effects analysis revealed that only ResNet101 showed a significant difference in the pareidolia effect between pretraining conditions (Bonferroni-corrected $p < 0.01$; other models, Bonferroni-corrected $p > 0.13$).

**Table 4.**

*Probability of categorizing face pareidolia images as faces in CNNs as compared with their false alarm baselines (i.e., categorizing regular object images as faces in the transfer task).*

| CNN | Mean (%) | Baseline (%) | Pareidolia effect (%, mean – baseline) | *t*(299) | *p* | Cohen's *d* |
|---|---|---|---|---|---|---|
| | | | Animal images included | | | |
| DenseNet169 | 47 | 38 | 9 | 6.9 | < .001 | 0.21 |
| ResNet101 | 24 | 11 | 13 | 9.10 | < .001 | 0.25 |
| VGG16 | 37 | 31 | 6 | 3.55 | < .001 | 0.27 |
| VGG13 | 37 | 31 | 6 | 4.06 | < .001 | 0.27 |
| VGG11 | 38 | 32 | 6 | 3.38 | < .001 | 0.29 |
| | | | Animal images excluded | | | |
| DenseNet169 | 52 | 41 | 11 | 9.34 | < .001 | 0.25 |
| ResNet101 | 29 | 24 | 5 | 3.47 | < .001 | 0.26 |
| VGG16 | 38 | 35 | 3 | 1.68 | 0.09 | 0.28 |
| VGG13 | 37 | 31 | 6 | 3.85 | < .001 | 0.28 |
| VGG11 | 37 | 35 | 2 | 1.45 | .15 | 0.28 |

Next, to further assess face-specific processing, the inversion effect was examined using previously "unseen" face images (from VGGFace2) and object images (from STL-10). The face inversion effect was evident across models, with all models except DenseNet169 showing reduced face categorization probabilities for inverted faces compared to upright ones (see **Table 5**; main effect of model type, $F(4, 2990) = 156.92$, $p < .001$, partial $\eta^2 = .17$). The presence of animal images in pretraining enhanced the inversion effect by 1.50% (main effect of training condition, $F(1, 2990) = 7.03$, $p = .008$, partial $\eta^2 = .002$), but the impact varied across models (interaction of training condition and model type, $F(4, 2990) = 11.33$, $p < .001$, partial $\eta^2 = .015$). Simple effects analysis revealed that only ResNet101's inversion effect was modulated by animal images in pretraining, with an increase of 8.7% ($F(1, 2990) = 49.67$, $p < .001$, partial $\eta^2 = .02$). This demonstrates that CNNs develop specialized face processing mechanisms similar to human perception, even when trained on simplified datasets without face or animal images.

On the other hand, object categorization performance remained stable despite inversion, regardless of the presence of animal images in pretraining datasets (see **Table 5**; main effect of training condition, $F(1, 2990) = 1.47$, $p = .23$), highlighting the face-specific nature of the inversion effect. Among all models, only VGG16's object inversion effect was modulated by the presence of animal images during pretraining, with a 3.9% stronger effect when animal images were included in pretraining ($F(1, 2990) = 9.36$, $p = .002$, partial $\eta^2 = .003$; interaction of training condition and model type, $F(4, 2990) = 2.99$, $p = .02$, partial $\eta^2 = .004$). When the object inversion effect served as the baseline and subtracted away from the face inversion effect, the same face inversion pattern still held (see **Table 5**).

**Table 5.**

*Face inversion and object inversion effects in CNNs.*

| CNN | Face inversion effect (%, upright – inverted) | *T* test | Object inversion effect (%, upright – inverted) | *T* test | Difference (%, face inversion – object inversion) | *T* test |
|---|---|---|---|---|---|---|
| | | | Animal images included | | | |
| VGG11 | **7.1** | **$t(531.50) = 5.55$, $p < .001$, $d = 0.16$** | 1.1 | $t(598) = 0.57$, $p = .57$, $d = 0.23$ | **6.0** | **$t(598) = 2.70$, $p = .01$, $d = 0.27$** |
| VGG13 | **7.5** | **$t(566.85) = 5.43$, $p < .001$, $d =$** | 1.9 | $t(598) = 1.05$, $p = .30$, $d = 0.23$ | **5.6** | **$t(598) = 2.39$, $p = .02$, $d = 0.28$** |
| | | **0.17** | | | | |
| VGG16 | 8.7 | ***t*(565.22) = 6.27, *p* < .001, *d* = 0.17** | **5.0** | ***t*(598) = 2.57, *p* = .01, *d* = 0.24** | 3.7 | *t*(598) = 1.58, *p* = .12, *d* = 0.28 |
| ResNet101 | **26.5** | ***t*(514.16) = 12.99, *p* < .001, *d* = 0.25** | -1.3 | *t*(577.35) = -1.13, *p* = .26, *d* = 0.14 | **27.8** | ***t*(561.55) = 12.13, *p* < .001, *d* = 0.28** |
| DenseNet169 | -0.3 | *t*(598) = -0.20, *p* = .84, *d* = 0.17 | -1.1 | *t*(598) = -0.79, *p* = .43, *d* = 0.17 | 0.8 | *t*(598) = 0.40, *p* = .69, *d* = 0.24 |
| | | | Animal images excluded | | | |
| VGG11 | **7.8** | ***t*(565.87) = 5.35, *p* < .001, *d* = 0.18** | -0.8 | *t*(598) = -0.43, *p* = .67, *d* = 0.22 | **8.7** | ***t*(598) = 3.78, *p* < .001, *d* = 0.28** |
| VGG13 | **6.3** | ***t*(572.01) = 4.89, *p* < .001, *d* = 0.16** | 3.4 | *t*(598) = 1.82, *p* = .07, *d* = 0.23 | 3.0 | *t*(598) = 1.32, *p* = .19, *d* = 0.28 |
| VGG16 | **8.9** | ***t*(567.93) = 6.05, *p* < .001, *d* = 0.18** | 1.2 | *t*(598) = 0.60, *p* = .55, *d* = 0.23 | **7.8** | ***t*(598) = 3.33, *p* < .001, *d* = 0.29** |
| ResNet101 | **17.6** | ***t*(541.95) = 9.49, *p* < .001, *d* = 0.23** | -1.6 | *t*(598) = -0.90, *p* = .37, *d* = 0.21 | **19.1** | ***t*(588.06) = 7.64, *p* < .001, *d* = 0.31** |
| DenseNet169 | 1.2 | *t*(598) = -1.06, *p* = .29, *d* = 0.14 | 0.03 | *t*(598) = 0.01, *p* = .99, *d* = 0.16 | 1.2 | *t*(598) = 0.72, *p* = .47, *d* = 0.21 |

*Note*. significant *p* values were marked in bold.

## Object and face processing units in the final FC layer

To test whether the two pretraining set configurations (with or without animal images) led to different distributions of processing units in the final FC layer after transfer learning, we next analyzed the proportions of face-specific units, object-specific units, and overlap units (units responsive to both face and object properties). As shown in Fig 7, there was no significant differences in the distribution of face, object, and overlap units between different pretrain datasets across the five CNN architectures (VGG11: $\chi^2$ = 6.0, $p$ = 0.20; VGG13: $\chi^2$ = 6.0, $p$ = 0.20; VGG16: $\chi^2$ = 6.0, $p$ = 0.20; ResNet101: $\chi^2$ = 6.0, $p$ = 0.20; DenseNet169: $\chi^2$ =

6.0, $p = 0.20$). The results confirm that the inclusion or exclusion of animal images during pretraining does not significantly influence the composition of processing units, demonstrating the robustness of CNNs in generating stable and specialized unit representations for faces and objects.

**Fig 7 | Proportions of activated units in the final FC layer of CNN models for face, object, and overlap categories.** Unit activations were recorded when exposed to face, object, and pareidolia images after transfer learning. (a) Proportions of face-active (blue), object-active (orange), and overlap (gray) units in models pretrained on datasets including animal images. (b) Proportions of face-active, object-active, and overlap units in models pretrained on datasets excluding animal images. Across all CNN architectures (VGG11, VGG13, VGG16, ResNet101, DenseNet169), the proportions of these unit types remained consistent, regardless of whether animal images were included or excluded during pretraining.

## Representational distance among faces, pareidolia, and objects

As in Experiment 1, we calculated the representational distances between face, pareidolia, and object images in the final FC layer of the CNNs. As Table 6 shows, despite variations in the representational distances across different units and models, the object–pareidolia distance was significantly shorter than the object–face distances. In addition, all models exhibited overlap units that effectively distinguished between face and object images, consistent with the findings from Experiment 1. Importantly, the inclusion or exclusion of animal images during pretraining did not significantly alter the capacity of overlap units to separate these categories, suggesting that these representations emerged robustly under both conditions.

**Table 6**

*Tensor distances between face, face pareidolia images, and objects in the last FC layer*

| CNN | Comparison | All units | | Overlap units | |
|---|---|---|---|---|---|
| | | 95% CIs | Distance | 95% CIs | Distance |
| | | Animal images included | | | |
| VGG11 | Face vs. pareidolia | [18.36, 20.47] | 19.25 | [11.40, 12.83] | 8.44* |
| | Object vs. pareidolia | [10.52, 12.29] | 11.04 | [6.12, 7.52] | 5.08* |
| | Face vs. object | [19.82, 21.90] | 20.71 | [13.79, 15.27] | 9.06* |
| VGG13 | Face vs. pareidolia | [20.28, 22.68] | 21.21 | [11.52, 13.17] | 8.67* |
| | Object vs. pareidolia | [14.71, 16.97] | 15.41 | [7.20, 8.53] | 6.14* |
| | Face vs. object | [21.15, 23.30] | 22.00 | [13.75, 15.30] | 8.94* |
| VGG16 | Face vs. pareidolia | [10.31, 11.50] | 10.79 | [5.61, 6.33] | 4.54* |
| | Object vs. pareidolia | [7.14, 8.30] | 7.52 | [3.30, 3.95] | 2.96* |
| | Face vs. object | [10.82, 11.95] | 11.28 | [6.38, 7.09] | 4.58* |
| ResNet101 | Face vs. pareidolia | [18.96, 20.67] | 19.70 | [8.99, 9.82] | 7.90* |
| | Object vs. pareidolia | [10.64, 12.01] | 11.03 | [4.29, 4.79] | 4.16* |
| | Face vs. object | [19.08, 20.56] | 19.71 | [9.55, 10.38] | 7.88* |
| DenseNet169 | Face vs. pareidolia | [1.95, 3.16] | 2.31 | [1.17, 1.91] | 1.53 |
| | Object vs. pareidolia | [1.84, 3.63] | 2.57 | [0.96, 1.90] | 1.75 |
| | Face vs. object | [2.17, 2.83] | 2.41 | [1.27, 1.56] | 1.83* |
| | | | Animal images excluded | | |
| VGG11 | Face vs. pareidolia | [11.92, 13.36] | 12.49 | [5.38, 6.14] | 5.02* |
| | Object vs. pareidolia | [8.64, 9.97] | 9.04 | [4.04, 4.12] | 3.86* |
| | Face vs. object | [12.62, 14.13] | 13.26 | [5.93, 6.67] | 5.41* |
| VGG13 | Face vs. pareidolia | [11.76, 13.24] | 12.37 | [7.03, 8.01] | 5.94* |
| | Object vs. pareidolia | [7.46, 8.69] | 7.82 | [4.02, 5.00] | 3.54* |
| | Face vs. object | [12.63, 14.19] | 13.29 | [8.49, 9.61] | 6.30* |
| VGG16 | Face vs. pareidolia | [9.19, 10.39] | 9.64 | [4.60, 5.21] | 4.53* |
| | Object vs. pareidolia | [7.24, 8.71] | 7.72 | [3.99, 5.16] | 3.63* |
| | Face vs. object | [10.17, 11.75] | 10.81 | [5.82, 6.98] | 5.22* |
| ResNet101 | Face vs. pareidolia | [16.35, 18.06] | 17.06 | [8.13, 9.02] | 7.26* |
| | Object vs. pareidolia | [9.47, 10.65] | 9.79 | [4.30, 4.96] | 4.21* |
| | Face vs. object | [17.30, 18.80] | 17.93 | [8.91, 9.73] | 7.80* |
| DenseNet169 | Face vs. pareidolia | [2.05, 3.46] | 2.48 | [1.27, 2.18] | 1.59 |
| | Object vs. pareidolia | [2.28, 4.41] | 3.19 | [1.30, 2.40] | 2.10 |
| | Face vs. object | [2.39, 3.20] | 2.72 | [1.58, 1.95] | 2.07* |

| | | Face units | | Object units | |
|---|---|---|---|---|---|
| | | 95% CIs | Distance | 95% CIs | Distance |
| | | | Animal images included | | |
| VGG11 | Face vs. pareidolia | [14.98, 16.82] | 11.64* | [13.35, 14.92] | 11.35* |
| | Object vs. pareidolia | [7.49, 8.89] | 7.06* | [7.77, 9.47] | 6.91* |
| | Face vs. object | [16.70, 18.51] | 12.403* | [15.67, 17.27] | 12.10* |
| VGG13 | Face vs. pareidolia | [15.86, 17.85] | 13.86* | [13.32, 15.00] | 12.12* |
| | Object vs. pareidolia | [9.89, 11.47] | 9.99 | [9.64, 11.41] | 8.73* |
| | Face vs. object | [16.97, 18.81] | 14.38* | [15.60, 17.36] | 12.66* |
| VGG16 | Face vs. pareidolia | [8.04, 9.05] | 6.59* | [6.89, 7.71] | 6.71* |
| | Object vs. pareidolia | [4.56, 5.29] | 4.33* | [5.11, 6.19] | 4.41* |
| | Face vs. object | [8.18, 9.11] | 6.84* | [7.95, 8.83] | 6.99* |
| ResNet101 | Face vs. pareidolia | [13.20, 14.45] | 12.37* | [11.40, 12.40] | 11.19* |
| | Object vs. pareidolia | [6.54, 7.32] | 6.69 | [6.39, 7.19] | 6.00* |
| | Face vs. object | [13.56, 14.69] | 12.31* | [12.00, 12.93] | 10.99* |
| DenseNet169 | Face vs. pareidolia | [1.44, 2.31] | 1.84 | [1.48, 2.36] | 1.79 |
| | Object vs. pareidolia | [1.34, 2.63] | 2.07 | [1.20, 2.32] | 2.03 |
| | Face vs. object | [1.60, 2.08] | 2.09* | [1.62, 2.00] | 2.07* |
| | | | Animal images excluded | | |
| VGG11 | Face vs. pareidolia | [8.61, 9.73] | 7.63* | [7.29, 8.16] | 7.25* |
| | Object vs. pareidolia | [6.09, 7.06] | 5.68* | [5.51, 6.40] | 5.29* |
| | Face vs. object | [9.42, 10.56] | 8.14* | [7.73, 8.61] | 7.68* |
| VGG13 | Face vs. pareidolia | [9.41, 10.70] | 8.29* | [8.27, 9.32] | 8.12* |
| | Object vs. pareidolia | [5.24, 6.22] | 4.94* | [5.34, 6.52] | 4.86* |
| | Face vs. object | [10.64, 12.00] | 9.00* | [9.59, 10.88] | 8.81* |
| VGG16 | Face vs. pareidolia | [5.98, 6.76] | 6.10 | [6.58, 7.41] | 6.59 |
| | Object vs. pareidolia | [4.80, 5.93] | 4.78* | [5.56, 7.07] | 5.16* |
| | Face vs. object | [6.95, 8.09] | 6.86* | [8.02, 9.62] | 7.28* |
| ResNet101 | Face vs. pareidolia | [11.81, 13.02] | 11.30 | [10.29, 11.38] | 10.75 |
| | Object vs. pareidolia | [6.54, 7.39] | 6.59 | [5.96, 6.79] | 6.15 |
| | Face vs. object | [12.66, 13.77] | 12.07* | [11.10, 12.06] | 11.47 |
| DenseNet169 | Face vs. pareidolia | [1.48, 2.44] | 1.88 | [1.61, 2.70] | 1.92 |
| | Object vs. pareidolia | [1.54, 2.89] | 2.46 | [1.59, 3.02] | 2.52 |
| | Face vs. object | [1.81, 2.32] | 2.29 | [1.87, 2.36] | 2.32 |

*Note.* CIs = 95% confident intervals; $*p < .05$

# Experiment 3

Experiment 2 corroborated our findings from Experiment 1, demonstrating that CNNs develop robust face processing capabilities independent of animal image presence during pretraining. In both experiments, CNNs were presented with 300 distinct face identities during inversion testing, and both unit-level correlation analysis and representational distance analysis suggested that overlapping units within these models may encode identity-specific facial features. Building on these findings, we evaluated whether these pretrained models could perform more complex face identity recognition tasks, by testing them on a large-scale identity recognition transfer task involving 1,000 distinct identity classes.

## Methods

We evaluated face identity recognition capabilities using the pretrained CNN models from Experiments 1 and 2 as base networks. For the transfer learning task, we used the VGGFace2 dataset [60], selecting 1,000 distinct identities for classification. Following the same transfer learning protocol used in Experiments 1 and 2, we froze all network parameters except those in the final FC layer, which were updated during training (batch size = 32, learning rate = 0.0001).

## Results

Models demonstrated robust performance on the face identity recognition task across all pretraining conditions (Fig 8). Models pretrained on ImageNet (Experiment 1) achieved the highest average validation accuracy at 34%, with DenseNet169 performing best (45%) and VGG16 lowest (26%). Models pretrained on Caltech-256 (both with and without animal images) reached an average validation accuracy of 20%, with ResNet101 showing the strongest performance (33%) and DenseNet169 the weakest (7%). All models significantly exceeded chance performance (0.1%) on this 1,000-class identification task.

**Figure 8 | Training and validation performance of five CNN models (VGG11, VGG13, VGG16, ResNet101, and DenseNet169) on face identity recognition across 40 epochs under different pretraining conditions.** Performance patterns were consistent across all pretraining conditions, with models showing similar

learning trajectories and final performance levels regardless of animal image presence during pretraining. All models demonstrated rapid initial learning, with performance metrics stabilizing within the first 10-15 epochs. DenseNet169 and ResNet101 consistently achieved higher accuracy levels compared to the VGG series.

# Discussion

Across three experiments, we investigated face processing mechanisms using five CNN architectures: three VGG variants (VGG11, VGG13, VGG16) and two contemporary models (DenseNet169, ResNet101). Despite being initialized only on object recognition (with object pretrain weights), these networks demonstrated remarkable face processing capabilities after transfer learning that targeted only the final FC layer. The models achieved near human-level performance in face-object categorization (Experiments 1&2), exhibited classic face-specific effects like the face pareidolia effect and inversion effect (Experiments 1&2), and successfully performed identity recognition tasks (Experiment 3). Analysis of the final FC layer revealed both specialized and overlapping units for face and object processing, with face representations forming distinct clusters separate from object representations in the activation space. In addition, by manipulating the training input (with versus without animal images), we show that the effect was robust. These findings support a domain-general account of face processing, suggesting that specialized face processing mechanisms can emerge from general object recognition architecture through learning.

The validity of using CNNs as models for human face processing depends on their ability to replicate key human behavioral patterns. Our inversion effect results provide evidence for such parallels: most CNN models demonstrated a face-specific inversion effect similar to humans, showing greater performance decrements for inverted faces compared to inverted objects. However, DenseNet169 in Experiment 2 (pretrained with Caltech-256) showed a notable deviation from this pattern, exhibiting no inversion effect for faces (or objects). This distinctive behavior may stem from DenseNet169's more sophisticated architecture, suggesting it might process visual information differently from simpler architectures. Previous research using LayerCAM methods [72] has shown that DenseNet169 extracts more facial features than VGG models, indicating potential differences in processing strategies between simpler and more advanced architectures. Future research employing techniques such as lesion studies or backward correlation analysis could further elucidate how processing mechanisms vary across model architectures of different complexities.

The face pareidolia results demonstrated broad similarities between CNN and human performance, while also revealing important nuances about training data composition. In Experiment 2, controlling for animal images in the pretraining dataset helped isolate the training requirements for face processing capabilities. However, this manipulation revealed a trade-off: while providing cleaner experimental conditions, removing entire image categories risks creating dataset imbalances that could impair model generalization [73,74]. This limitation became apparent when VGG11 and VGG16 failed to show significant pareidolia effects after animal image removal, suggesting reduced robustness in detecting face-like structures in novel contexts. Nevertheless, the overall patterns observed in Experiment 2 largely replicated those from Experiment 1, indicating that the presence of animal images in pretraining data is not critical for developing face processing capabilities.

These findings validate previous studies that removed only human faces from training data, while suggesting that the standard ImageNet-1K database remains a practical choice for pretraining, balancing experimental control against the risks of dataset imbalance. Indeed, [75] found significant differences in alignment accuracies between animal and human faces using identical keypoint numbers, highlighting their fundamental structural differences. This categorical distinction is well-established in cognitive neuroscience, as exemplified by neuroimaging studies treating faces and animals as separate stimulus categories [76].

CNNs were more likely than humans to categorize pareidolia images as objects. This disparity may, in part, reflect differences in task framing—CNNs were tasked with determining whether an image depicted a face or an object, whereas human participants assessed whether the image contained face-like patterns. Additionally, humans have undergone vastly more extensive “training,” having encountered a much broader variety of stimuli throughout their lifetimes compared to CNNs. Another key distinction lies in the role of top-down processing, which is integral to human face perception in pareidolia [45,77–81]. Humans perceive configurations of objects that resemble faces as closely akin to real human faces [38,82,83]. Even minimal cues can trigger false face perceptions, such as recognizing animal faces [70,84,85], a phenomenon that likely underlies face detection in pareidolia images [86,87].

Despite these differences in task structure, lifelong exposure to diverse stimuli, and the absence of top-down processing in CNNs, the models exhibited qualitatively similar patterns of pareidolia perception to humans. This similarity is notable, given their general lack of robustness to distortions [88–91]. Future research could explore strategies such as incorporating blurred images during training to enhance CNN robustness and refine their ability to mimic human perceptual processes [92,93].

The reduced effectiveness of CNNs in recognizing pareidolia images aligns with the challenges observed in individuals with autism [94–97] and schizophrenia [98–100]. In addition, similar to autism [101]; for a review, see [102], all five CNNs exhibited a weakened face inversion effect compared to typical human effects. These patterns suggest a potential parallel between cognitive processing in these individuals and the algorithmic strategies employed by CNNs. These CNN models may therefore be used as potential computational models to study the visual processing of neuroatypical populations.

Our findings appear to contrast with previous studies supporting domain-specific face processing in CNNs [41,103]. This discrepancy likely stems from crucial methodological differences. Previous approaches typically analyzed activation patterns from object-trained CNNs directly exposed to face images, without first establishing their capacity for face processing [41,42]. This is akin to recording electrophysiological activity from a visual cortex neuron in response to auditory inputs and interpreting the activity as evidence of auditory processing. The activation patterns observed in these prior studies may reflect generic activation rather than true “face processing activation,” thereby failing to capture the mechanisms underlying face perception.

To address this limitation, we employed transfer learning in Experiments 1 and 2 to ensure our networks could effectively process facial information before analyzing their representations. This ensured that the CNNs were explicitly trained to process face images. Notably, we identified overlap units that reliably represented and differentiated between face and object images. These overlap units resemble discrete face-selective regions interspersed among other functionally specific cortical areas in primates [104,105]. Moreover, the distinct activation patterns elicited by different stimuli within these units align with the multifunctionality characteristic of neuronal processing [106]. These findings provide compelling evidence in support of the domain-general hypothesis, highlighting the adaptability of object recognition systems in processing a wide range of stimuli, including faces.

The representational richness of the pretraining datasets also affect model capacity [107]. Experiment 3 demonstrated that the model pretrained on ImageNet-1K significantly outperformed the model pretrained on the smaller and less diverse Caltech-256 dataset in recognizing celebrity identities. While previous studies often used weights trained on specific categories or subordinate-level objects, such as “cars” [103] or “birds” [108], our approach leveraged weights trained to distinguish among a thousand object categories. This broader pretraining provided the models with richer representations, enhancing their capacity to process a variety of stimuli. Thus, even when CNNs can encode faces using feature spaces derived from object representations, insufficient representational richness may hinder transfer task performance. This limitation may explain why previous studies often updated multiple network layers during transfer learning [103,109], whereas we modified only the final FC layer to ensure face-specific features emerged from object recognition architectures. Furthermore, our study uniquely combined both subordinate-level classification (distinguishing specific identities) and superordinate-level categorization (face versus object categorization), better reflecting natural visual processing hierarchies [55].

In conclusion, our findings demonstrate that CNNs pretrained exclusively on object recognition can develop robust face processing capabilities with minimal additional training. Despite having no initial exposure to faces, these networks exhibited key signatures of human face perception, including face-specific inversion effects and pareidolia phenomena. These results provide compelling evidence that sophisticated face processing, including identity recognition, can emerge from domain-general object recognition mechanisms rather than requiring face-specific neural architectures. Moreover, the similarities between CNN performance patterns and those observed in neurodivergent populations suggest that these models could serve as valuable computational frameworks for studying atypical face processing mechanisms.

## Constraints on Generality

Our findings draw strength from the use of diverse, globally sourced datasets of faces and face pareidolia images, as well as established CNN architectures that demonstrate human-like perceptual mechanisms [33–35]. However, important limitations exist. While CNNs replicated many aspects of human face processing, they showed reduced sensitivity to face pareidolia compared to human observers. This difference likely reflects the absence of top-down processing mechanisms in current CNN architectures—a crucial component of human pareidolic face perception [77,78,110]. Notably, these limitations parallel patterns observed in certain clinical populations, as our CNNs demonstrated similarities to individuals with autism and schizophrenia in both pareidolia perception and face inversion effects. This alignment suggests potential applications of CNN models in studying the computational mechanisms underlying atypical face processing.

## Authors' Contributions

**Z. Zhao:** Coding and Design, Data Analysis, Data Visualization, and Writing (Initial Draft and Final Draft). **J. Chen:** Data Analysis, Writing (Suggestion and Final Draft). **Z. Lin:** Design, Data Analysis, Interpretation, Data Visualization, Writing (Initial Draft and Final Draft). **H. Ying:** Conceptualization, Design, Data Analysis, Data Visualization, Funding & Resources Acquisition, Supervision, and Writing (Initial Draft and Final Draft).

## Acknowledgement

We thank Shangzhi Lu for assistance in creating Figures 3 and 4 using R. We thank Dr Benjamin Baker for constructive comments and suggestions. The authors declared no conflict of interest. H. Y. is supported by the National Natural Science Foundation of China (32200850).